\UseRawInputEncoding
\documentclass[runningheads,a4paper]{llncs}
\usepackage[top=2.5cm, bottom=2.5cm, left=2.5cm, right=2.5cm]{geometry}
\usepackage{amsmath}
\usepackage{amssymb}
\usepackage{graphicx}
\usepackage{booktabs}
\usepackage{listings}
\usepackage{xcolor}
\usepackage{hyperref}
\usepackage{caption}
\usepackage{subcaption}

\usepackage{url}
\usepackage{color}
\usepackage{cite}
\usepackage{epsfig}
\usepackage{soul}

\usepackage[nomargin,inline,marginclue,draft]{fixme}
\usepackage{cleveref}
\fxsetup{status=draft}
\fxsetup{theme=color, mode=multiuser} \FXRegisterAuthor{me}{ame}{\color{red}Me}
\newcommand{\orcid}[1]{\href{https://orcid.org/#1}{\includegraphics[scale=0.02]{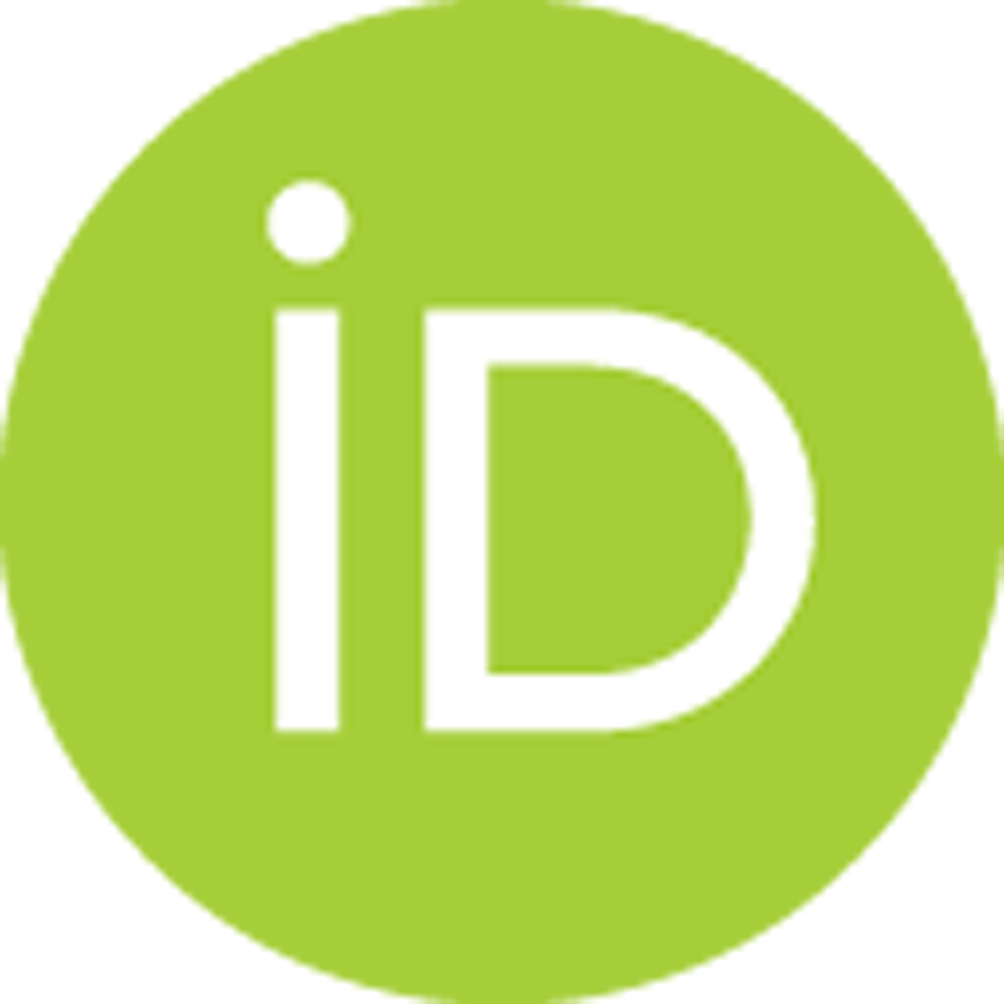}}} % this does not print the whole orcid number

\hypersetup{
    colorlinks=true
}
\title{BraTS-Path Challenge: Assessing Heterogeneous Histopathologic Brain Tumor Sub-regions}
\titlerunning{BraTS-Path Challenge}

\author{
Spyridon Bakas\inst{1,2,3,4,5,6,\dag,\S,\P,\orcid{0000-0001-8734-6482}}%checked
\and Siddhesh P. Thakur\inst{1,\dag,\S,\orcid{0000-0003-4807-2495}}%checked
\and Shahriar Faghani\inst{7,\dag,\S,\orcid{0000-0003-3275-2971}}%checked
\and Mana Moassefi\inst{7,\dag,\orcid{0000-0002-0111-7791}}%checked
\and Ujjwal Baid\inst{1,3,\dag,\orcid{0000-0001-5246-2088}}%checked
\and Verena Chung\inst{8,\dag,\orcid{0000-0002-5622-7998}}%checked
\and Sarthak Pati\inst{1,6,\dag,\orcid{0000-0003-2243-8487}}%checked
\and Shubham Innani\inst{1,\dag,\orcid{0000-0003-3616-0308}}%checked
\and Bhakti Baheti\inst{1,\dag,\orcid{0000-0001-5475-3903}}%checked
\and Jake Albrecht\inst{8,\dag,\orcid{0000-0003-1813-7649}}%checked
\and Alexandros Karargyris\inst{6,\dag,\orcid{0000-0002-1930-3410}}
\and Hasan Kassem\inst{6,\dag,\orcid{0000-0001-5830-8890}}
\and MacLean P. Nasrallah\inst{9,\dag,\orcid{0000-0003-4861-0898)}}
\and Jared T. Ahrendsen\inst{10,\dag,\S,\orcid{0000-0002-9309-6544}}%checked
\and Valeria Barresi\inst{11,\dag,\S,\orcid{0000-0001-7086-1920}}%checked
\and Maria A. Gubbiotti\inst{12,\dag,\S,\orcid{0000-0003-2246-7958}}%checked
\and Giselle Y. López\inst{13,14,15,\dag,\S,\orcid{0000-0001-5435-6668}}%checked
\and Calixto-Hope G. Lucas\inst{16,\dag,\S,\orcid{0000-0002-8347-9592}}%checked
\and Michael L. Miller\inst{17,\dag,\S,\orcid{0000-0002-6350-5706}}%checked
\and Lee A. D. Cooper\inst{10,\dag,\S,\orcid{0000-0002-3504-4965}}%checked
\and Jason T. Huse\inst{12,18\dag,\S,\orcid{0000-0003-4514-0640}}%checked
\and William R. Bell\inst{1,\dag,\S}
}

\authorrunning{S.Bakas, et al.}

\institute{\scriptsize{
Department of Pathology and Laboratory Medicine, Indiana University School of Medicine, Indianapolis, IN, USA%1
\and
Department of Neurological Surgery, Indiana University School of Medicine, Indianapolis, IN, USA%2
\and
Department of Radiology and Imaging Sciences, Indiana University School of Medicine, Indianapolis, IN, USA%3
\and
Department of Biostatistics and Health Data Science, Indiana University School of Medicine, Indianapolis, IN, USA%4
\and
Department of Computer Science, Luddy School of Informatics, Computing, and Engineering, Indiana University, Indianapolis, IN, USA%5
\and Medical Research Group, MLCommons, San Francisco, CA, USA%6
\and Department of Radiology, Mayo Clinic, MN, USA%7
\and Sage Bionetworks, USA%8
\and Department of Pathology and Laboratory Medicine, Perelman School of Medicine, University of Pennsylvania, Philadelphia, PA, USA%9
\and Department of Pathology, Feinberg School of Medicine, Northwestern University, Chicago, IL, USA%10
\and Department of Diagnostics and Public Health, University of Verona, Verona, Italy%11
\and Division of Pathology and Laboratory Medicine, Department of Anatomical Pathology, University of Texas MD Anderson Cancer Center, TX, USA%12
\and The Preston Robert Tisch Brain Tumor Center at Duke, Durham, NC, USA%13
\and Department of Pathology, Duke University School of Medicine, Durham, NC, USA%14
\and Department of Neurosurgery, Duke University Medical School, Durham, NC, USA%15
\and Division of Neuropathology, Department of Pathology, Johns Hopkins University School Of Medicine, Baltimore, MD, USA%16
\and Department of Pathology and Cell Biology, Columbia University Irving Medical Center, New York, NY, USA%17
\and Department of Translational Molecular Pathology, University of Texas MD Anderson Cancer Center, TX, USA%18
}
\linebreak
\\
\textsuperscript{\dag} People involved in the organization of the challenge.\\
\textsuperscript{\S} People involved in annotation process.\\ 
\textsuperscript{\P} Corresponding author: \email{\{spbakas@iu.edu\}}}

\begin{document}
    \mainmatter
    \maketitle
    \setcounter{footnote}{0}
    \begin{abstract}
        Glioblastoma is the most common primary adult brain tumor, with a grim prognosis - median survival of 12-18 months following treatment, and 4 months otherwise. Glioblastoma is widely infiltrative in the cerebral hemispheres and well-defined by heterogeneous molecular and micro-environmental histopathologic profiles, which pose a major obstacle in treatment. Correctly diagnosing these tumors and assessing their heterogeneity is crucial for choosing the precise treatment and potentially enhancing patient survival rates. In the gold-standard histopathology-based approach to tumor diagnosis, detecting various morpho-pathological features of distinct histology throughout digitized tissue sections is crucial. Such ``features'' include the presence of cellular tumor, geographic necrosis, pseudopalisading necrosis, areas abundant in microvascular proliferation, infiltration into the cortex, wide extension in subcortical white matter, leptomeningeal infiltration, regions dense with macrophages, and the presence of perivascular or scattered lymphocytes. With these features in mind and building upon the main aim of the BraTS Cluster of Challenges \url{https://www.synapse.org/brats2024}, the goal of the BraTS-Path challenge is to provide a systematically prepared comprehensive dataset and a benchmarking environment to develop and fairly compare deep-learning models capable of identifying tumor sub-regions of distinct histologic profile. These models aim to further our understanding of the disease and assist in the diagnosis and grading of conditions in a consistent manner.
    \end{abstract}
    
    
    % =========================
    % main part of the document
    % =========================
    \section{Introduction}

    The International Brain Tumor Segmentation (BraTS) challenge describes a landmark benchmarking environment and a dataset for the fair evaluation and development of artificial intelligence (AI) algorithms for identifying distinct histologic glioma sub-regions in neuro-pathology digitized tissue sections. More specifically, since 2012, the BraTS challenge has focused on benchmarks and datasets associated with the distinction and quantification of brain tumor sub-regions that numerous studies have shown association with morbidity and mortality \cite{menze2014multimodal, bakas2017advancing, bakas2017segmentation_1, bakas2017segmentation_2,bakas2022university, baid2021rsna,bakas2018identifying,baid2020novel}. However, all BraTS challenges to-date have been focusing on the domain of radiology and particularly on multi-parametric Magnetic Resonance Imaging (mpMRI) scans, which define the front-line assessment for diagnosis, treatment planning, and longitudinal treatment monitoring by radiologists, neurosurgeons, neuro-oncologists, and radiation oncologists; yet our understanding of the histological sub-regions of brain tumors is limited to qualitative evaluations.
    
    This year the BraTS Cluster of Challenges partners with the AI-RANO (Artificial Intelligence for Response Assessment in Neuro Oncology) group to present newly proposed clinically relevant challenges, in a synergistic attempt to maximize the potential clinical impact of the innovative algorithmic contributions made by the international participating teams. Although the focus still remains on the generation of a common benchmarking environment, it also has a further expanded 1) clinical relevance, 2) scope, and 3) datasets, by introducing an independent challenge with a particular focus on the distinct histopathology tumor sub-regions by assessing histology samples from digitized tissue sections of brain glioma.

    This paper introduces the BraTS-Path challenge, which seeks the assessment of the heterogeneous histologic landscape of glioma by automatically detecting and quantifying different morphological regions in whole slide histopathology images (WSI) of brain glioblastoma (GBM) (WHO Gr.4, IDH-wt). The BraTS-Path challenge will provide a community standard and benchmark for state-of-the-art automated classification of tissue patches from a WSI based on a deep expert annotated multilabel dataset. Challenge competitors will develop automated classification models to predict distinct sub-regions on WSI, including cellular tumor, pseudopalisading necrosis, areas abundant in microvascular proliferation, geographic necrosis, infiltration into the cortex, penetration into white matter, leptomeningeal infiltration, regions dense with macrophages, and presence of lymphocytes. AI models will be evaluated on separate validation and independent hold-out test datasets, using standardized metrics \cite{maier2024metrics,reinke2024understanding}. The models developed during this challenge will contribute in moving automated brain tumor classification algorithms a step closer to clinical practice, towards furthering our understanding of this disease and ultimately improve brain tumor patient care.
    
    \section{Materials \& Methods}
        \subsection{Data Description}

    In the BraTS-Path 2024 challenge dataset, we focus on glioblastoma (GBM) digitized tissue sections with representative features. The H\&E-stained, Formalin-fixed, paraffin-embedded (FFPE) digitized tissue sections used in the BraTS-Path challenge, describe histology images acquired during standard clinical practice across 11 International sites. The exact staining process details and the digital scanners (with their technical specifications) used for acquiring this cohort are not publicly available neither through The Cancer Imaging Archive (TCIA) \cite{clark2013cancer}, nor through the Genomic Data Commons (GDC) Data Portal of the NIH/NCI. The acquisition protocols and equipment are different across (and within each) contributing institution, as these represent real routine clinical practice.

    The TCGA-GBM and TCGA-LGG data sets, which are publicly accessible via the TCIA, have been chosen for this challenge. Initially, we have reclassified these collections in line with the 2021 WHO classification of CNS tumors. This reclassification was specifically done to pinpoint all cases of GBM IDH-wildtype, which are categorized under CNS WHO grade 4. The TCGA-LGG collection, initially classified as low-grade astrocytomas, is redefined under the 2021 WHO CNS criteria as GBMs due to specific molecular characteristics indicative of distinct tumor evolution. Consequently, these astrocytomas, now classified as molecular GBMs, are included in this challenge to develop algorithms applicable to all clinical GBMs per WHO guidelines. Conversely, certain cases in the TCGA-GBM collection have been excluded because their molecular profiles do not align with the current WHO definition of GBM. For this study, a single H\&E-stained tissue section from each case in the reclassified TCGA-GBM and TCGA-LGG collections is used. Focusing solely on FFPE slides, we avoid hydration artifacts common in frozen sections.

\subsection{Participating sites}  

    The provided data have been acquired from:
    \begin{enumerate}
        \item Henry Ford Hospital (MI, USA),
        \item University of California (CA, USA),
        \item MD Anderson Cancer Center (TX, USA),
        \item Emory University (GA, USA),
        \item Mayo Clinic (MN, USA),
        \item Thomas Jefferson University (PA, USA),
        \item Duke University School of Medicine (NC, USA),
        \item Saint Joseph Hospital and Medical Center (AZ, USA),
        \item Case Western Reserve University (OH, USA),
        \item University of North Carolina (NC, USA),
        \item Fondazione IRCCS Instituto Neuroligico C. Besta, (Italy),
    \end{enumerate}
    Note that data from these institutions are provided through The Cancer Imaging Archive (TCIA -http://www.cancerimagingarchive.net/), supported by the Cancer Imaging Program (CIP) of the National Cancer Institute (NCI) of the National Institutes of Health (NIH).

\subsection{Annotation Protocol}
    \begin{figure*}[h]
        \centering
        \includegraphics[width=1\linewidth]{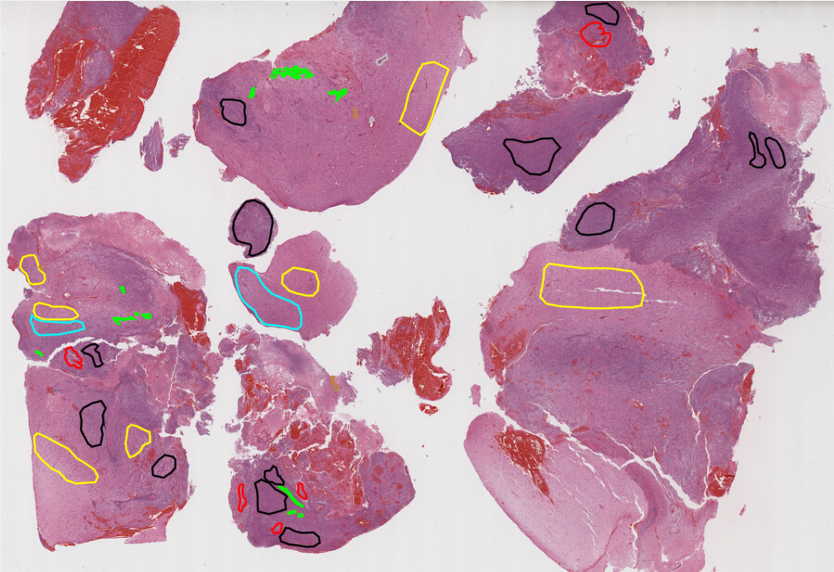}
        \caption{GBM Pathology sub-regions considered from one of the WSI in the BraTS-Path 2024 Challenge. The image considers various regions of GBM including Pseudopalasading Necrosis (Red), Microvascular Proliferation (Green), Necrosis (Blue), Infiltration into the cortex (Yellow), Cellular Tumor (Black), and Penetration into white matter (Sky blue). Other regions such as the Presence of Lymphocytes (PL), Regions Dense with Macrophages (DM), and Leptomeningeal Infiltration (LI) are not present in the given slide and are not annotated but are available through other slides.}
        \label{fig:enter-label}
    \end{figure*}

    The annotation of these data followed a specific pre-defined clinically-approved annotation protocol (defined by expert neuropathologists), which was provided to all clinical annotators, describing in detail instructions on what the segmentations of each histologic feature should describe (see below for the summary of the specific instructions). The annotators were given the flexibility to use their tool of preference for making the annotations, or the provided infrastructure based on the Digital Slide Archive (DSA) \cite{gutman2017digital} available through a web portal by the Division of Computational Pathology of the Department Pathology and Laboratory Medicine at Indiana University School of Medicine (IUSM), and follow a complete manual annotation approach.
    
    Summary of specific histologic areas of interest:
    \begin{enumerate}
        \item presence of cellular tumor (CT)
        \item pseudopalisading necrosis (PN)
        \item areas abundant in microvascular proliferation (MP)
        \item geographic necrosis (NC)
        \item infiltration into the cortex (IC)
        \item penetration into white matter (WM)
        \item leptomeningeal infiltration (LI)
        \item regions dense with macrophages (DM)
        \item presence of lymphocytes (PL)
    \end{enumerate}

    Each case was assigned to a pair consisting of an annotator and an approver. Furthermore, a select number (3 exactly) of cases were annotated by all annotators to conduct a separate assessment of inter-rater variability and determine acceptable variations in the automated solutions. Annotators spanned across various experience levels and clinical/academic ranks, while the approver was an experienced board-certified neuropathologist (with $>$10 years of experience). The annotators were given the flexibility to use their tool of preference for making the annotations, or the provided infrastructure based on the IUSM instance of DSA (\url{https://dsa.sca.iu.edu/dsa/}), and follow a complete manual annotation approach. Once the annotators were satisfied with the produced annotations, they passed these to the corresponding approver. The approver is then responsible for signing off these annotations. Specifically, the approver would review the tumor annotations, in tandem with the corresponding tissue section, and the annotations of not satisfactory quality were removed from the provided annotation. If the patches from the remaining annotations were less than approximately 1,500 patches then the tissue sections would be sent back to the annotators for further annotations. This iterative approach was followed for all cases until their respective annotations reached satisfactory quality (according to the approver) for being publicly available and noted as final ground truth segmentation labels for these cases. Subsequently, these regions were segmented into patches classified based on the presence of specific histology. This approach established a classification task aimed at accurately identifying patches with specific morphologic features.

\subsection{Data Processing}

    Based on the given definition that a case encompasses data processed to produce one result that is compared to the corresponding reference result, a case in this challenge represents an individual patch extracted from an H\&E-stained, FFPE digitized tissue section of a single patient tumor at a specific timepoint. We ensured that the patches were of a similar size, with each representing either a specific class present in that patch or none, in which case it was classified as `background'.

    These tissue sections exhibit a variety of features indicative of the diagnosis of a GBM and have been annotated by expert neuropathologists. These annotated regions are divided into same-size patches, each of them corresponding to a distinct morpho-histologic feature (or class) that the participants are expected to predict. Since this task has not been conducted before, we consider individual patches as individual cases in this challenge, intending to conduct a detailed analysis and offer a deeper understanding of these distinct features/classes in a more fine-grained resolution. Specifically, we would like to assess the intrinsic similarity of these classes and hence inherent difficulty of detecting individual classes, as well as which are the most confused with each other features/classes.

    These patches are classified according to their respective features/classes throughout the training, validation, and testing phases. The presence of histologic features characteristic of GBM determined the inclusion criteria for each tissue section. Please note that all tissue sections included for each case of the provided dataset, represent the tissue sections with the best quality available for this particular case.

    Post-annotation of histologically distinct regions by clinical experts, each region is segmented into 512 x 512 patches. This specific patch size was chosen to minimize noise at the boundaries which could occur due to smaller patch sizes and ensure that the patches fully cover the annotated regions. Since the annotations were made following careful consideration of the WSIs, avoiding tissue folding, pen markings, and glass slippage, additional rigorous patch-level curation was not required. However, if measures are not taken to ensure annotations are made on clean regions, rigorous quality control could become a critical step to avoid affecting downstream tasks (see Discussion for more details).
    
    The challenge participants can obtain the labeled training data at any point from the Synapse platform (\url{https://www.synapse.org/brats}). These data will be used to develop, containerize, and evaluate their algorithms in unseen validation data until July 2024, when the organizers will stop accepting new submissions and evaluate the submitted algorithms in the hidden testing data. Ground truth reference annotations for all datasets are created and approved by expert neuropathologists for every subject included in the training, validation, and testing datasets to evaluate the performance of the participating algorithms quantitatively. 

\subsection{Defining Distinct Sub-regions}

    \subsubsection{Cellular Tumor (CT):} This sub-region is a crucial histopathological characteristic defining tumor cellularity, playing a significant role in differentiating lower-grade gliomas (LGGs) from higher-grade gliomas (HGGs), and is closely associated with tumor aggressiveness and patient prognosis \cite{Bui2024}. LGGs typically exhibit lower cell density than HGGs, significantly influencing diagnostic radiology imaging features, such as apparent diffusion coefficient (ADC) and magnetic resonance imaging (MRI) values. In contrast, the high cellularity observed in HGGs is strongly correlated with rapid tumor growth and poorer patient outcomes. This distinct difference in cellularity is essential in guiding the selection of appropriate therapeutic strategies and predicting patient survival rates, making it a key factor in the management of gliomas \cite{Bulakba2019}.

    \begin{figure*}[h]
        \centering
        \includegraphics[width=0.5\linewidth]{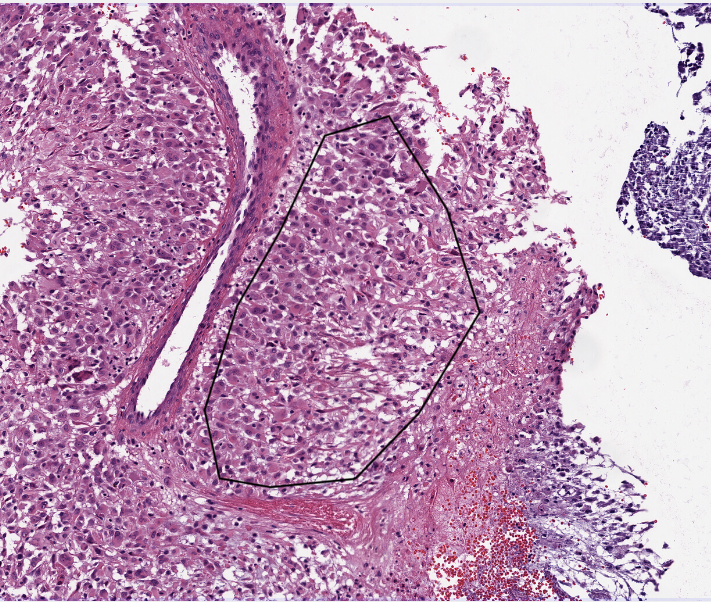}
        \caption{Sample region of interest of selected tissue for Cellular Tumor}
        \label{fig:CT}
    \end{figure*}

    \subsubsection{Pseudopalisading Necrosis (PN):} This is a sub-compartment often observed in HGGs \cite{Ludwig2021}, and especially glioblastomas (GBM, WHO Gr.4, IDH-wt). It is characterized by rows of tumor cells surrounding necrotic areas and is associated with increased tumor grade and aggressive behavior, indicating a hypoxic environment and rapid tumor growth \cite{Park2022}. It's an important diagnostic marker in neuroimaging and neuropathology, significantly affecting prognosis and treatment decisions. The presence of PN typically suggests a worse prognosis due to the tumor's aggressive nature and its association with other malignant features like vascular proliferation.

\begin{figure*}[ht]
    \centering
    \begin{subfigure}[b]{0.485\linewidth}
        \centering
        \includegraphics[width=\linewidth]{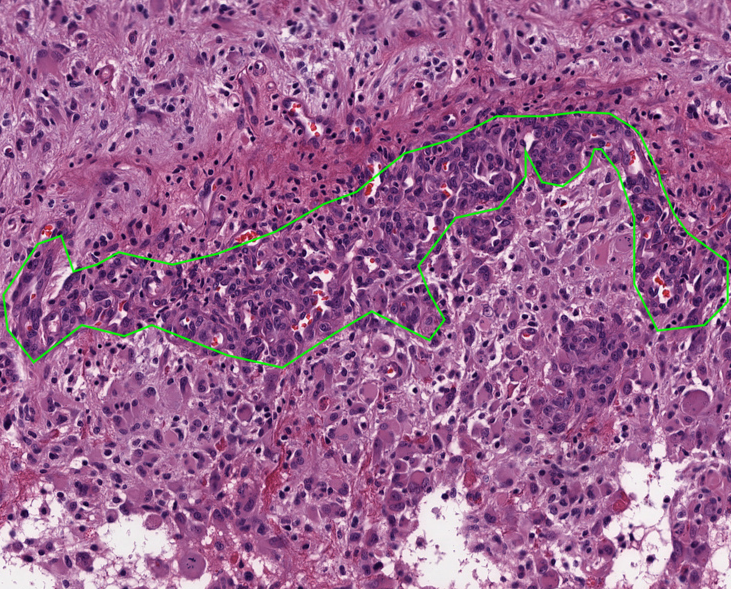}
        \caption{Microvascular Proliferation}
        \label{fig:PSN}
    \end{subfigure}%
    \hfill
    \begin{subfigure}[b]{0.485\linewidth}
        \centering
        \includegraphics[width=\linewidth]{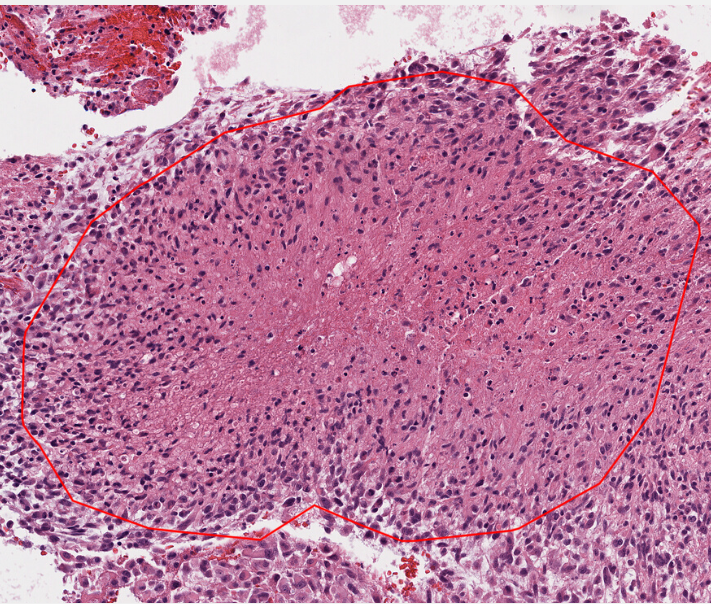}
        \caption{Pseudopalisading Necrosis}
        \label{fig:MP}
    \end{subfigure}
    \vskip\baselineskip
    \begin{subfigure}[b]{0.485\linewidth}
        \centering
        \includegraphics[width=\linewidth]{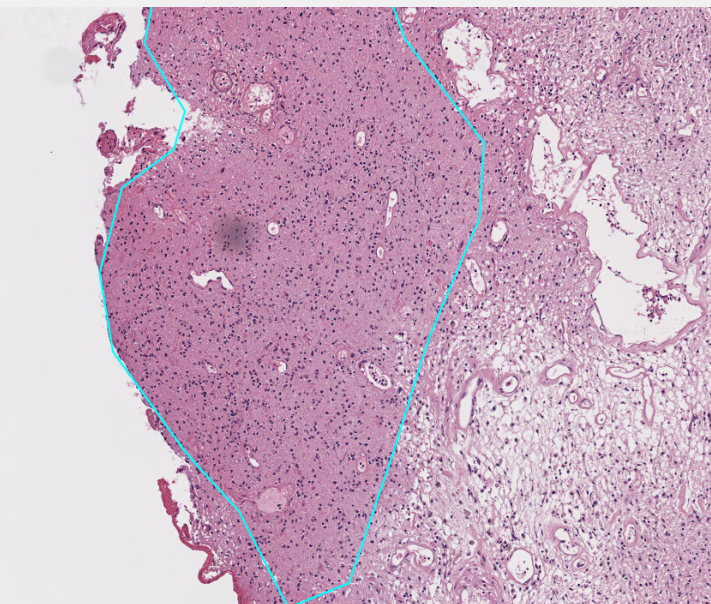}
        \caption{White Matter}
        \label{fig:PWM}
    \end{subfigure}%
    \hfill
    \begin{subfigure}[b]{0.485\linewidth}
        \centering
        \includegraphics[width=\linewidth]{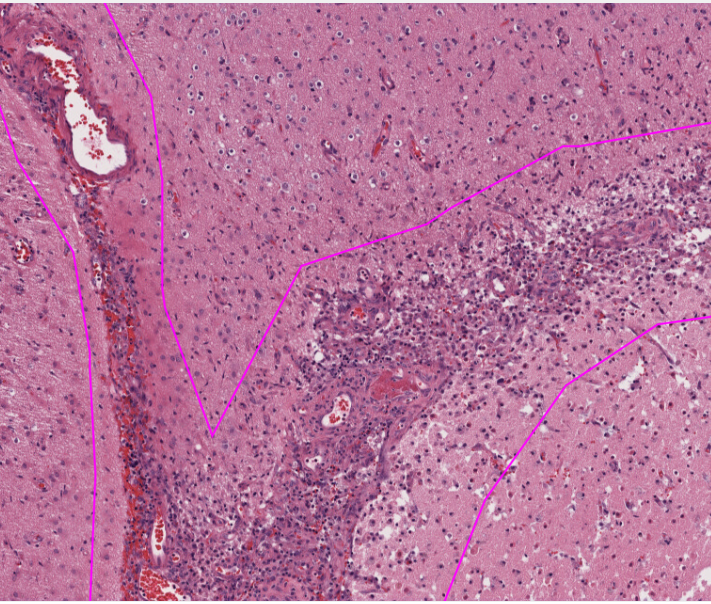}
        \caption{Leptomeningeal Infiltration}
        \label{fig:LI}
    \end{subfigure}
    \caption{Sample regions of interest of selected tissues. (a)Microvascular Proliferation, (b)  Pseudopalisading Necrosis,  (c) White Matter, (d) Leptomeningeal Infiltration.}
    \label{fig:combined}
\end{figure*}

    \subsubsection{Microvascular Proliferation (MP):} This feature is more commonly observed in HGGs and involves the formation of new, abnormal blood vessels within the tumor, driven by the tumor's increased metabolic demands and hypoxia \cite{Stadlbauer2020}. This feature is a critical diagnostic criterion distinguishing between LGGs and HGGs and is linked to tumor grade and aggressiveness. In radiology imaging studies, microvascular proliferation results in higher relative cerebral blood volume (rCBV) on perfusion MRI, aiding in the differentiation and grading of gliomas.

    \subsubsection{Penetration into White Matter (WM):} The invasive nature of gliomas, particularly HGGs, includes extensive infiltration into surrounding white matter that can be detected by diffusion tensor MRI, which shows reduced integrity of white matter tracts\cite{Zhang2022, DSouza2019}. This characteristic complicates complete surgical resection and is a prognostic factor for increased recurrence and reduced overall survival. These penetrations can also be observed in the extent of white matter invasion, which is crucial for planning treatment approaches and assessing potential neurological deficits post-treatment.

    \subsubsection{Leptomeningeal Infiltration (LI):} Leptomeningeal infiltration in gliomas, especially in HGGs, involves spreading cancer cells into the leptomeninges, the membranes covering the brain and spinal cord \cite{Park2022}. This condition is associated with advanced disease stages and significantly worsens the prognosis. Detection of leptomeningeal spread is important for staging the tumor and planning appropriate interventions, such as radiation therapy or chemotherapy.
    
    \subsubsection{Tumor Necrosis (TN):} This feature is commonly seen in HGGs, indicative of rapid tumor growth and high metabolic activity, and is associated with poor prognosis\cite{Homma2006}. The extent of necrosis can influence treatment decisions, as it reflects the aggressive nature of the tumor and potential resistance to certain therapies. In clinical practice, the assessment of necrosis helps in the grading of gliomas and can guide the aggressiveness of the therapeutic approach.

    \subsubsection{Regions of Dense Macrophages (DM):} These are often found in glioma, surrounding necrotic areas, where they play roles in phagocytosing cell debris\cite{Mller2017}. In HGGs, abundant macrophages are linked with inflammation and tumor progression, impacting the tumor's microenvironment and patient prognosis. This feature is also used to assess the immune response within the tumor and guide immunotherapy strategies.
    
\begin{figure*}[h]
    \centering
    \begin{subfigure}[b]{0.485\linewidth}
        \centering
        \includegraphics[width=\linewidth]{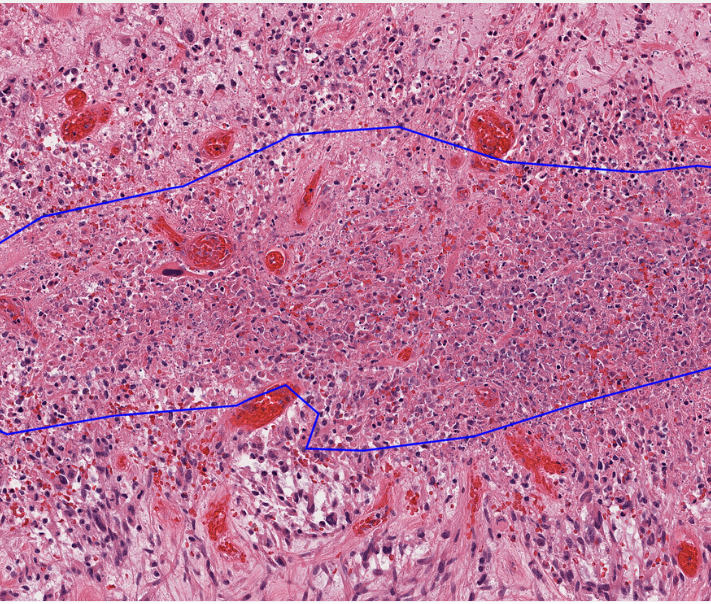}
        \caption{Tumor Necrosis}
        \label{fig:TN}
    \end{subfigure}%
    \hfill
    \begin{subfigure}[b]{0.485\linewidth}
        \centering
        \includegraphics[width=\linewidth]{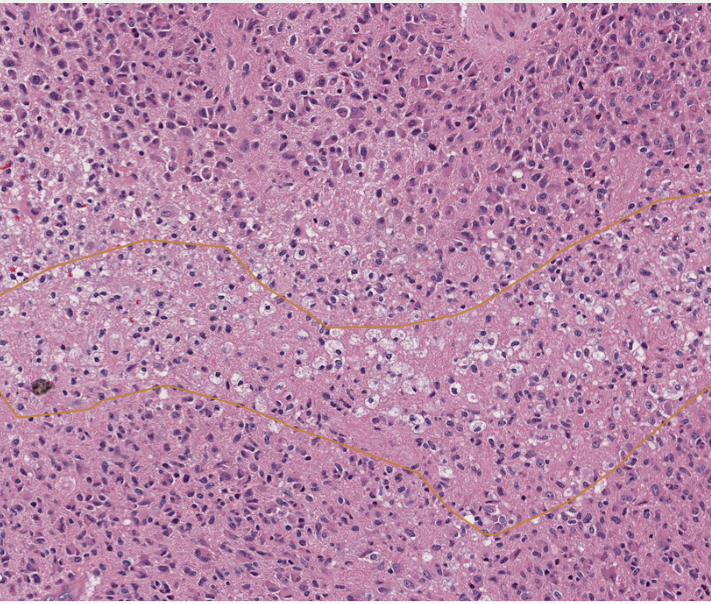}
        \caption{Regions with Dense Macrophages}
        \label{fig:RDM}
    \end{subfigure}
    \vskip\baselineskip
    \begin{subfigure}[b]{0.485\linewidth}
        \centering
        \includegraphics[width=\linewidth]{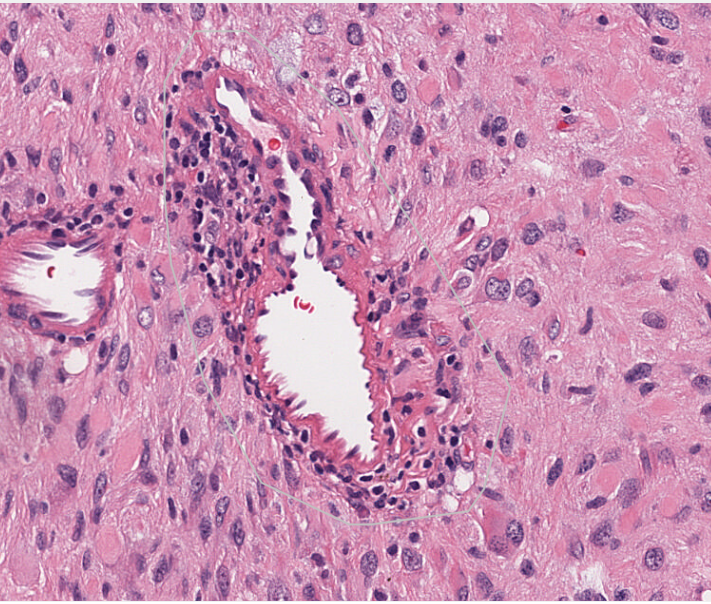}
        \caption{Presence of Lymphocytes}
        \label{fig:PL}
    \end{subfigure}%
    \hfill
    \begin{subfigure}[b]{0.485\linewidth}
        \centering
        \includegraphics[width=\linewidth]{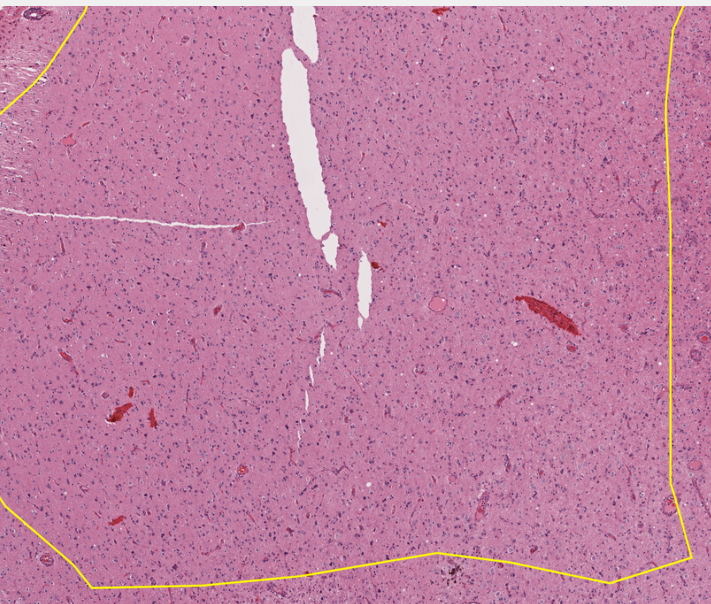}
        \caption{Infiltrating Cortex}
        \label{fig:IC}
    \end{subfigure}
    \caption{Sample regions of interest of selected tissues. (a) Tumor Necrosis, (b) Regions with Dense Macrophages, (c) Presence of Lymphocytes, (d) Infiltrating Cortex.}
    \label{fig:combined}
\end{figure*}

    \subsubsection{Presence of Lymphocytes (PL):} Lymphocytic infiltration in gliomas varies with tumor grade \cite{Rutledge2013}. It is more pronounced in LGGs and correlates with a better immune response and potentially better prognosis. The presence of lymphocytes can indicate the body's immune defense against tumor cells, and exploring this feature could help develop immunotherapeutic treatments.
    
    \subsubsection{Infiltration into Cortex (IC):} Infiltration of gliomas into the cortical regions significantly impacts clinical outcomes, as it often correlates with higher-grade tumors and more severe neurological symptoms \cite{DAlessio2019}. The degree of cortical invasion can guide surgical planning and influence decisions regarding the extent of resection and the use of adjunctive therapies like chemotherapy or radiation.

    \subsection{Data Distribution}
\begin{figure}[ht]
            \centering
%            \begin{subfigure}[b]{0.5\linewidth}
                \centering
                \includegraphics[width=0.6\linewidth]{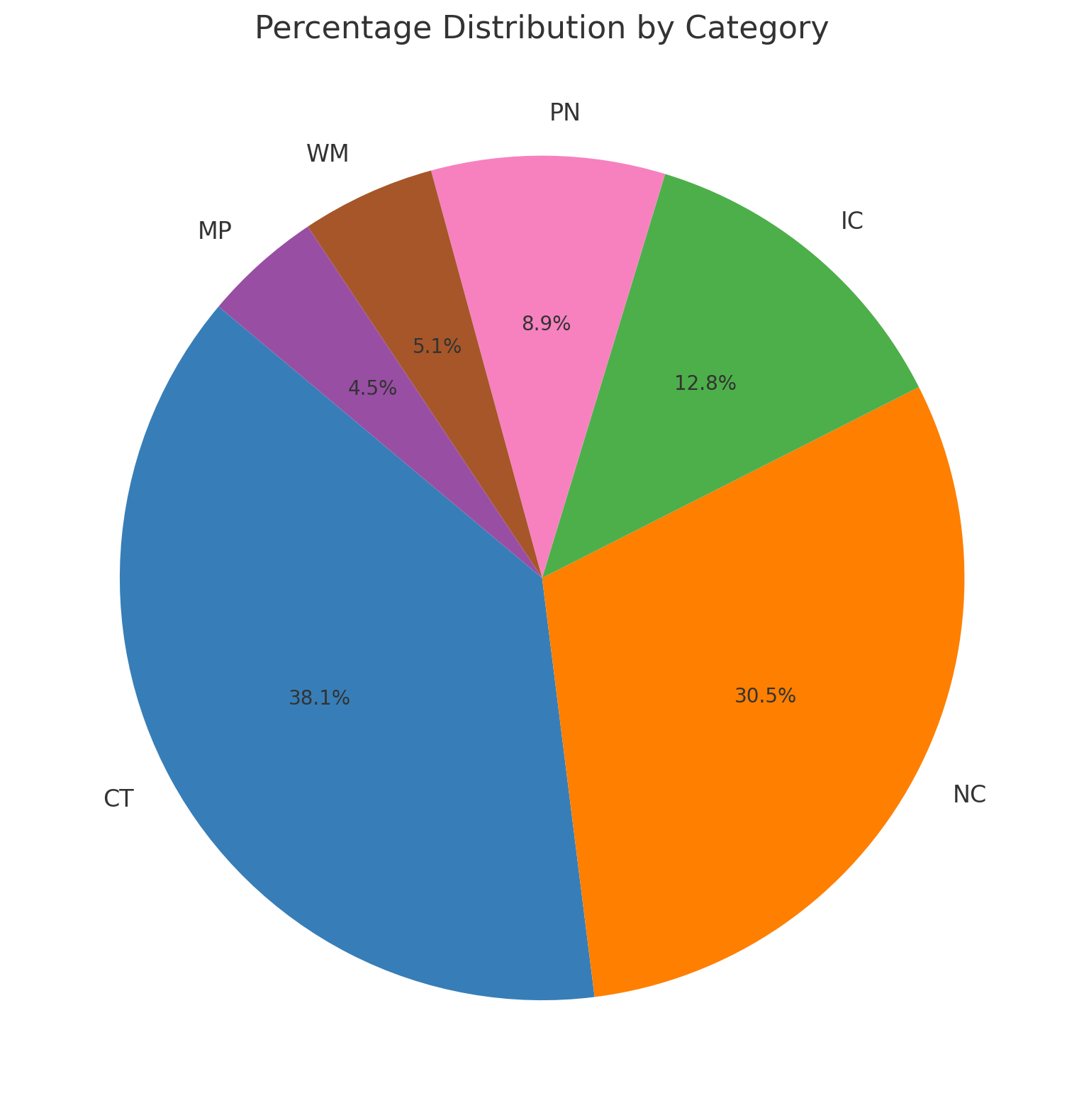}
                \caption{Pie chart showing the percentage distribution of values of the number of patches by region of interest included in the challenge.}
                \label{fig:subfig1}
%            \end{subfigure}%
%            \begin{subfigure}[b]{0.5\linewidth}
%                \centering
%                \includegraphics[width=\linewidth]{figures/bargraph.png}
%                \caption{Bar graph displaying the values for each category. The bars are color-coded according to the categories.}
%                \label{fig:subfig2}
%            \end{subfigure}
            \label{fig:combined}
        \end{figure}

        After rigorous annotations and patch extraction, the distribution in Fig.\ref{fig:subfig1} was obtained. Since the class imbalance is high, we focus on the six regions with the largest number of available patches in this year's challenge. In Table \ref{tab:values_roi_patches}, we have underlined the selected regions of interest, considering the class imbalance and the challenge it presents to participants.
        \begin{table}[ht]
            \centering
            \begin{tabular}{|c|c|}
                \hline
                \textbf{ROI} & \textbf{Number of Patches} \\
                \hline
                \underline{\textbf{CT}}  & \underline{\textbf{43401}} \\
                \underline{\textbf{NC}}  & \underline{\textbf{24438}} \\
                \underline{\textbf{IC}}  & \underline{\textbf{14500}} \\
                \underline{\textbf{PN}} & \underline{\textbf{10101}} \\
                \underline{\textbf{WM}} & \underline{\textbf{5791}} \\
                \underline{\textbf{MP}} & \underline{\textbf{5115}} \\
                DM & 1788  \\
                LI  & 1534 \\
                PL & 672 \\
                \hline
            \end{tabular}
            \caption{Values, ROI, and Number of Patches for Categories}
            \label{tab:values_roi_patches}
        \end{table}
        
\subsection{Evaluation Assessment}

    Evaluation of submissions is planned to be performed based on containerized approaches using the MLCommons MedPerf platform. MedPerf is an open federated AI/ML evaluation platform \cite{karargyris2023federated} that is meant to automate the evaluating pipeline by running the participants’ models on the testing datasets of each contributing site’s data and calculating evaluation metrics on the resulting predictions. Finally, the Synapse platform (\url{https://www.synapse.org/brats}) will retrieve the metrics results from the MedPerf server and will rank them to determine the top-ranked methods.
    
    In terms of evaluation metrics, we use:

    Accuracy: This fundamental metric will provide the proportion of true results (both true positives and true negatives) among the total number of cases examined. It proves the effectiveness of the classification model.

    \textbf{F1 Score}: As a measure that balances precision (positive predictive value) and recall (sensitivity), the F1 score is crucial for scenarios where the cost of false positives and false negatives is high. It is particularly useful when dealing with imbalances between classes.
    \begin{equation*}
        \textbf{F1 Score} = 2 \times \frac{\text{Precision} \times \text{Recall}}{\text{Precision} + \text{Recall}}
    \end{equation*}

    \textbf{Matthews Correlation Coefficient (MCC)}: MCC provides a balanced measure even when the classes are of very different sizes. It is a correlation coefficient between the observed and predicted classifications, offering a more informative and nuanced assessment than simple accuracy.
    \begin{equation*}
        \textbf{MCC} = \frac{TP \times TN - FP \times FN}{\sqrt{(TP + FP)(TP + FN)(TN + FP)(TN + FN)}}
    \end{equation*}
    where TP is True Positive, TN is True Negative, FP is False positive, and FN is False Negative.

    \textbf{Area under the Receiver Operator Characteristic Curve (AUROC)}: This curve quantifies the overall ability of the model to discriminate between the positive and negative classes across different thresholds. A higher AUROC indicates better model performance

    \begin{equation*}
        \textbf{AUROC} = \int_{0}^{1} TPR(FPR) \, d(FPR)
    \end{equation*}
    where TPR is the True Positive Rate (Sensitivity), and FPR is the False Positive Rate.

    \textbf{Sensitivity and Specificity}: to determine whether an algorithm tends to over- or underclassify different classes.
    \begin{equation*}
        \textbf{Sensitivity (True Positive Rate)}: \text{Sensitivity} = \frac{TP}{TP + FN}
    \end{equation*}
    \begin{equation*}
        \textbf{Specificity (True Negative Rate)}: \text{Specificity} = \frac{TN}{TN + FP}
    \end{equation*}
    where TP is True Positive, TN is True Negative, FP is False Positive, and FN is False Negative.

    All of them will be included in the open-source metrics calculation package using the Generally Nuanced Deep Learning Framework (GaNDLF) (\url{gandlf.org}) \cite{pati2023gandlf}.

        \section{Discussion}

% \begin{itemize}
%     \item First challenge of this such type.
%     \item Patch level classification can be used for unsupervised segmentation of different intra-tumor parts of WSI.
% \end{itemize}
%\subsection{Potential benefits of the challenge}
    The BraTS-Path challenge offers a standardized benchmarking environment to develop automated AI models using the largest expert-annotated multilabel histopathology dataset of GBM to date with the potential to bridge the gap between research and clinical practice. By incorporating a diverse range of digitized FFPE tissue sections from multiple sites, the challenge dataset captures the heterogeneity inherent to GBM. This diversity introduces variability in the training dataset, reflecting real-world clinical scenarios and accounting for differences in tissue processing and imaging acquisition protocols across institutions. By providing a standardized benchmarking environment and a diverse training dataset, the challenge fosters the development of highly generalizable segmentation models. Furthermore, the collaborative nature of the BraTS-Path challenge fosters a community-driven approach to advancing the field of neuro-oncology. By bringing together researchers, neuropathologists, and data scientists from around the world, the challenge encourages knowledge sharing, collaboration, and interdisciplinary research. This collective effort not only accelerates progress in automated glioma histologic sub-region classification but also promotes the dissemination of best practices and standards within the research community.

    In the BraTS-Path 2024 challenge, annotations were made on carefully reviewed WSI, avoiding issues such as tissue folding, pen markings, and glass slippage, so additional rigorous patch-level curation was not required. However, if the annotations were not made on clean regions, rigorous patch-level image curation would be essential to distinguish between tissue-occupied areas and artifacts, such as glass reflection, pen markings, or tissue tearing/folding. This process has shown previously \cite{bahetiFrontiers2024} to obtain maximal benefits when including three distinct steps:
    \begin{enumerate}
        \item Removal of patches with significant white (intensity $>$230) or black (intensity $<$25) background, based on Red-Green-Blue (RGB) values, if such colors exceed 60\% of the patch.
        \item Conversion to Hue-Saturation-Value (HSV) space to eliminate patches with substantial saturation or value anomalies, discarding those where the percentage of such pixels exceeds 95\%.
        \item Stain deconvolution into Hematoxylin-Eosin-DAB (HED) space, discarding patches where over 80\% of Eosin channel pixels have an intensity below 50. These thresholds have been empirically determined to ensure that only artifacts are removed, preserving tissue-occupied areas in the selected patches.
    \end{enumerate}
    
% \begin{itemize}
%     \item Class-wise distribution patches in the complete dataset can be included in the data section. If there is class imbalance, that can be added as a limitation.
%     \item Comment on inter-annotator variability
%     \item There might be regions that are thin (less than patch size 256x256). Comment on how are they handled
%     \item labor- and time-intensive process of annotation
%     \item patch based task as compared to WSI
% \end{itemize}
%\subsection{Limitations of the challenge}
    While the BraTS-Path challenge represents a significant advancement in automated brain tumor sub-region classification, several limitations warrant consideration. Firstly, the class-wise distribution of patches within the dataset exhibits an imbalance, potentially impacting the performance of classification models. Patch imbalance could lead to biased predictions and hinder the generalizability of the algorithms. Additionally, inter-annotator variability among neuropathologists may introduce inconsistencies in the ground truth annotations, affecting the reliability of the dataset - However, we will be conducting an extensive analysis to understand this variability. Moreover, the presence of thin regions within digitized tissue sections, smaller than the designated patch size of 512 $\times$ 512, poses a challenge for algorithms, requiring robust handling techniques to ensure accurate classification. The patches were curated to ensure that a small region of interest (ROI) of 128 $\times$ 128 was captured within the larger 512 $\times$ 512 patch. Consequently, this patch would be labeled as the ROI. This approach was particularly used for PL and MP regions. Furthermore, the labor- and time-intensive process of manual annotation is a significant bottleneck in dataset creation, limiting the scalability of the challenge and potentially constraining the diversity and size of the dataset cohort. Lastly, the patch-based nature of the task, as opposed to WSI, may overlook contextual information crucial for accurate tumor classification, highlighting the need for future iterations of the challenge to address this limitation and transition toward WSI-level segmentation approaches.

%\subsection{Goals for future challenges}
    The BraTS-Path challenge aims to continue advancing the field of automated brain tumor classification by addressing emerging challenges and opportunities in neuro-oncology. Future iterations of the challenge will focus on refining and enhancing existing AI models. We will be adding denser annotations with multi-institutional data, increasing the number of classes/labels assessed, refining the dataset, and using the current dataset to predict tumor WHO grade, which would provide substantial clinical utility in determining prognosis and optimal treatment without requiring new imaging data. These advancements would significantly improve pathologist workflows and enhance objectivity in longitudinal follow-up assessments, contributing to more precise and effective patient care.

\section{Conclusion}

The BraTS-Path challenge represents a significant step forward in the field of neuro-oncology by addressing the critical need for accurate and automated identification of distinct histologic sub-regions within brain tumors, furthering our disease understanding. The dataset would enable the development and evaluation of AI models on a large dataset of digitized tissue sections providing valuable insights into the heterogeneity of GBM. By utilizing a vast dataset of expert-annotated histopathology images, the challenge establishes a robust benchmark for AI models, which would potentially contribute to clinical applications, including precise tumor monitoring, treatment planning, and non-invasive assessments of tumor characteristics. The challenge overarching goals are i) the transformative potential of AI in revolutionizing brain tumor care, paving the way for enhanced outcomes worldwide, and ii) fostering collaboration and knowledge sharing among researchers and clinicians, with the ultimate goal of accelerating the translation of research findings into tangible benefits for patients with brain tumors.
    
    % =========================
    \iffalse
    \section*{Author Contributions}
        Study conception and design: SB, JH, WRB
        Software development used in the study: SPT, UB, VC 
        Wrote the paper: SB, SPT, SI, BB
        Data analysis and interpretation: SB, SPT, JA, VB, MG, GYL, CHGL, DMM, MLM, JH, WRB
        Reviewed / edited the paper: All authors
    \fi
    
    \section*{Acknowledgments}
        %Developing large and well curated mpMRI datasets for auto-segmentation model development requires significant time and expertise from neuro-radiology experts. We are grateful to everyone who contributed to the development and review of the tumor volume labels including volunteer annotators/approvers from the American Society of Neuroradiology (ASNR).
         %This section should be added at the conclusion of the challenge, to avoid revealing data sources during the challenge. The data contributors are: 1)Center 1  2) Center 2 
    
    \section*{Funding}
    
    Research reported in this publication was partly supported by the Informatics Technology for Cancer Research (ITCR) funding program of the National Cancer Institute (NCI) of the National Institutes of Health (NIH) under award numbers U01CA242871, U24CA279629, and U24CA248265. The content of this publication is solely the responsibility of the authors and does not represent the official views of the NIH.
        
    \bibliographystyle{ieeetr}
    \bibliography{bibliography.bib}
    \newpage
    \appendix
\end{document}